\def\year{2019}\relax
\documentclass[letterpaper]{article} 
\usepackage{aaai19}  
\usepackage{times}  
\usepackage{helvet}  
\usepackage{courier}  
\usepackage{url}  
\usepackage{graphicx}  
\usepackage{amsfonts}
\usepackage{amssymb}
\usepackage{xspace}
\usepackage{booktabs}
\usepackage{algorithm}
\usepackage{algorithmic}
\usepackage{amsmath,bm}
\usepackage{dsfont}
\usepackage{multirow}
\usepackage [autostyle, english = american]{csquotes}

\MakeOuterQuote{"}
\usepackage{ntheorem,lipsum} 
\theorembodyfont{\upshape} 
\newtheorem{definition}{Definition}
\newtheorem{problem}{Problem}
\usepackage[colorlinks=true, allcolors=blue]{hyperref}

\newcommand{\mname}{\texttt{GAMENet}\xspace}
\begin{document}
%
\title{\mname : Graph Augmented MEmory Networks for Recommending Medication Combination}
\author{
Junyuan Shang$^{1,2}$, Cao Xiao$^3$, Tengfei Ma$^3$, Hongyan Li$^{1,2}$, Jimeng Sun$^4$ \\ 
\normalsize $^1$ Key Laboratory of Machine Perception, Ministry of Education, Beijing, China \\
\normalsize $^2$ School of EECS, Peking University, Beijing, China \\
\normalsize $^3$ IBM Research \\
\normalsize $^4$ Georgia Institute of Technology \\
\small sjy1203@pku.edu.cn, cxiao@us.ibm.com, Tengfei.Ma1@ibm.com, lihy@cis.pku.edu.cn, jsun@cc.gatech.edu
}
\maketitle
\begin{abstract}
Recent progress in deep learning is revolutionizing the healthcare domain including providing solutions to medication recommendations, especially recommending medication combination for patients with complex health conditions. Existing approaches either do not customize based on patient health history, or ignore existing knowledge on drug-drug interactions (DDI) that might lead to adverse outcomes. To fill this gap, we propose the Graph Augmented Memory Networks (\mname), which integrates the drug-drug interactions knowledge graph by a memory module implemented as a graph convolutional networks, and models  longitudinal patient records as the query. It is trained end-to-end to provide safe and personalized recommendation of medication combination. 
We demonstrate the effectiveness and safety of \mname by comparing with several state-of-the-art methods on real EHR data. \mname outperformed all baselines in all effectiveness measures, and also achieved $3.60\%$ DDI rate reduction from existing EHR data.

\end{abstract}

\section{Introduction}

 Today abundant health data such as longitudinal electronic health records (EHR) enables researchers and doctors to build better computational models for recommending accurate diagnoses and  effective treatments. Medication recommendation algorithms have been developed to assist doctors in making effective and safe medication prescriptions. 
A series of deep learning methods have been designed for medication recommendation. There are mainly two types of such methods: 1) \textit{Instance-based medication recommendation} models that perform recommendation based only on the current encounter and do not consider the longitudinal patient history, see ~\cite{zhang2017leap,wang2017safe}.
As a result, a patient with newly diagnosed hypertension will likely be treated the same as another patient who has suffered chronic uncontrolled hypertension.  Such a limitation affects accuracy and utility of the recommendations. 2) \textit{Longitudinal medication recommendation} methods such as ~\cite{choi2016retain,choi2016doctor,lipton2015learning,le2018dual} that leverage the temporal dependencies within longitudinal patient history to predict future medication. However, to our best knowledge, none of them considers drug safety in their modeling, especially ignoring the adverse drug-drug interactions (DDI) which are harder to prevent than single drug adverse reaction. Drugs may interact when they are prescribed and taken together, thus DDIs are common among patients with complex health conditions. Preventing DDIs is important since they could lead to health deterioration or even death.

To fill the gap, we propose Graph Augmented Memory Networks (\mname), an end-to-end deep learning model that takes both longitudinal patient EHR data and drug knowledge base on DDIs as inputs and aims to generate effective and safe recommendation of medication combination. In particular, \mname  consists of 
1) patient queries based on representations learned by a dual recurrent neural networks (Dual-RNN), and 2) an integrative and dynamic graph augmented memory module. 
It builds and fuses across multiple data sources (drug usage information from EHR and DDI knowledge from drug knowledge base~\cite{Tatonetti2012}) with graph convolutional networks (GCN)~\cite{kipf2016semi} in Memory Bank (\textbf{MB}).  The knowledge of combined uses of medications and drug-drug interaction relations are thus integrated.  It further writes patient history to dynamic memory (\textbf{DM}) in key-value form, which mimics case-based retrievals in clinical practice, i.e., considering similar patient representations from the DM.  Information from the graph augmented memory module can be retrieved by patient representation as query to generate memory outputs. Then, memory outputs and query will be concatenated to make effective and safe recommendations. \mname is optimized to balance between effectiveness and safety by combining multi-label prediction loss from EHR data and DDI loss for DDI knowledge. 

To summarize,  our work has the following contributions: 
\begin{itemize}
\item We jointly model the longitudinal patient records as an EHR graph and drug knowledge base as a DDI graph in order to provide  effective and safe medication recommendations. This is achieved by optimizing a combined loss that balances between multi-label prediction loss (for effectiveness) and DDI loss (for safety).
\item We propose graph augmented memory networks which embed multiple knowledge graphs using a late-fusion mechanism based GCN into the memory component  and enable attention-based memory search using query generated from longitudinal patient records. 
\item We demonstrate the effectiveness and safety of our model by comparing with several state-of-the-art methods on real EHR data. \mname outperformed all baselines in effectiveness measures, and achieved $3.60\%$ DDI rate reduction from existing EHR data (i.e., identify and reduce existing DDI cases compared with raw EHR data).
\end{itemize}

\section{Related Works}
\subsubsection{Memory Augmented Neural Networks} (MANN) have shown initial successes in NLP research areas such as question answering ~\cite{DBLP:journals/corr/WestonCB14,sukhbaatar2015end,miller2016key,kumar2016ask}. Memory Networks ~\cite{DBLP:journals/corr/WestonCB14} and Differentiable Neural Computers (DNC)~\cite{graves2016hybrid} proposed to use external memory components to assist the deep neural networks in remembering and storing things. After that, various MANN based models have been proposed such as ~\cite{sukhbaatar2015end,kumar2016ask,miller2016key}. In healthcare, memory networks can be valuable due to their capacities in memorizing medical knowledge and patient history. DMNC~\cite{le2018dual} proposed a MANN model for medication combination recommendation task using EHR data alone. In this paper, we use memory component to fuse multi-model graphs as memory bank to facilitate recommendation.

\subsubsection{Graph Convolutional Networks (GCN)} emerged for inducing informative latent feature representations of nodes from arbitrary graphs~\cite{kipf2016semi,DBLP:journals/corr/DefferrardBV16,DBLP:journals/corr/HamiltonYL17,chen2018fastgcn}. GCN models learn node embeddings in the following manner: Given each graph node initially attached with a feature vector, the embedding vector of each node are the transformed weighted sum of the feature vectors of its neighbors. All nodes are simultaneously updated to perform a layer of forward propagation. The deeper the network, the larger the local neighborhood. Thus global information is disseminated to each graph node for learning better node embeddings. GCNs haven been successfully used to model biomedical n etworks such as drug-drug interaction (DDI) graphs.  For example, ~\cite{DBLP:journals/corr/abs-1804-10850} models each drug as a node and DDIs as node labels in the drug association network and extended the GCN to embed multi-view drug features and edges. \cite{Zitnik2018} used GCN to model the drug interaction problems by constructing a large two-layer multimodal drug interaction graphs. In this paper, we use GCN to model medication as nodes and DDIs as links. 

\subsubsection{Medication Combination Recommendation}
could be categorized into instance-based and longitudinal medication recommendation methods. Instance-based methods focus on current health conditions. Among them, Leap~\cite{zhang2017leap} formulates a multi-instance multi-label learning framework and proposes a variant of sequence-to-sequence model based on content-attention mechanism to predict combination of medicines given patient's diagnoses. Longitudinal-based methods leverage the temporal dependencies among clinical events, see ~\cite{choi2016retain,choi2016doctor,lipton2015learning,le2018dual,doi:10.1093/jamia/ocy068}. Among them, RETAIN~\cite{choi2016retain} is based on a two-level neural attention model which detects influential past visits and significant clinical variables within those visits. DMNC~\cite{le2018dual} highlighted the memory component to enhance the memory ability of recurrent neural networks and combined DNC with RNN encoder-decoder to predict medicines based on patient's history records which has shown high accuracy. However, safety issue is often ignored by longitudinal-based methods. In this work, we design a memory component but target at building a structured graph augmented memory, where we not only embed DDI knowledge but also design a DDI loss to reduce DDI rate.

\section{Method}

\subsection{Problem Formulation}
\begin{definition}[Patient Records]
In longitudinal EHR data, each patient can be represented as a sequence of multivariate observations: $ \bm{P}^{(n)} = [ \bm{x}_1^{(n)}, \bm{x}_2^{(n)}, \cdots, \bm{x}_{T^{(n)}}^{(n)} ] $ where $n\in \{1,2,\ldots, N\}$, $N$ is the total number of patients; $T^{(n)}$ is the number of visits of the $n^{th}$ patient. To reduce clutter, we will describe the algorithms for a single patient and  drop the superscript $(n)$ whenever it is unambiguous. Each visit $\bm{x}_t = [\bm{c}_d^t, \bm{c}_p^t, \bm{c}_m^t]$ of a patient is concatenation of corresponding diagnoses codes $\bm{c}_d^t$, procedure codes $\bm{c}_p^t$ and medications codes $\bm{c}_m^t$.  For simplicity, we use $\bm{c}_\ast^t$ to indicate the unified definition for different type of medical codes. $\bm{c}_\ast^t \in \{0,1\}^{|\mathcal{C}_\ast|}$ is a multi-hot vector, where $\mathcal{C}_\ast$ denotes the medical code set and $|\mathcal{C}_\ast|$ the size of the code set.
\end{definition}

\begin{definition}[EHR\&DDI Graph]
EHR graph and DDI graph can be denoted as $G_e=\{\mathcal{V},\mathcal{E}_e\}$ and $G_d=\{\mathcal{V},\mathcal{E}_d\}$ respectively, where node set $\mathcal{V} = \mathcal{C}_m = \{c_{m_1}, c_{m_2},\cdots,c_{m_n}\}$ represents the set of medications, $\mathcal{E}_e$ is the edge set of known combination medication in EHR database and $\mathcal{E}_d$ is the edge set of known DDIs between a pair of drugs. Adjacency matrix $\bm{A}_e, \bm{A}_d \in \mathbb{R}^{|\mathcal{C}_m| \times |\mathcal{C}_m|}$ are defined to clarify the construction of edge $\mathcal{E}_e, \mathcal{E}_d$. For $\bm{A}_e$, we firstly create a bipartite graph with drug on one side and drug combination on the other side. Then $\bm{A}_e = \bm{A}_b \bm{A}_b^\intercal$ where $\bm{A}_b \in \mathbb{R}^{|\mathcal{C}_m| \times l}$ 
 is the adjacency matrix of the bipartite graph, $\bm{A}_b[i,j]=1$ when $i^{th}$ medication exists in $j^{th}$ medications combination and the number of unique medications combination denotes as $l$. For $\bm{A}_d$, only pair-wise drug-drug interactions are considered, $\bm{A}_d[i,j]=1$ when the $i^{th}$ medication has interaction with the $j^{th}$ one.

\end{definition}

\begin{problem}[Medication Combination Recommendation]
Given medical codes of the current visit at time $t$ (excluding medication codes) $\bm{c}_d^t, \bm{c}_p^t$, patient history $\bm{P} = [\bm{x}_1, \bm{x}_2, \cdots, \bm{x}_{t-1}]$ and EHR graph $G_e$, and DDI graph $G_d$, 
we want to recommend multiple medications by generating multi-label output $\hat{\bm{y}}_t \in \{0,1\}^{|\mathcal{C}_m|}$.

\end{problem}

\begin{table}[tb]
\centering
\caption{Notations used in \mname}
\label{tab:notation}
\begin{tabular}{l|l}
\toprule[1pt]
\bf Notation      & \bf Description                                            \\ \midrule
$\bm{P}\in \mathbb{R}^{|\mathcal{C}|\times T}$      & patient records    \\  
$\mathcal{C}_\ast$    & medical codes set of type $\ast$              					\\ 
$c_{\ast,i}$         & $i^{th}$ medical code in $\mathcal{C}_\ast$ of type $\ast$                 \\ 
$\bm{c}_\ast \in \mathbb{R}^{|\mathcal{C}_\ast|}$         & multi-hot vector of type $\ast$ \\
$\bm{x}_t \in \mathbb{R}^{|\mathcal{C}|}$    & concatenation of medical codes $\bm{c}_\ast^t$              					\\ \hline
$G_\ast$           & EHR or DDI Graph $\{\mathcal{V},\mathcal{E}_\ast\}$                \\ 
$\mathcal{V}$ &  vertex set same as $\mathcal{C}_m$            \\ 
$\mathcal{E}_\ast$ & edge set of $\ast$ dataset                                          \\ \hline
$\bm{e}_{\ast} \in \mathbb{R}^{d} $         & medical embeddings of type $\ast$               \\
$\bm{h}_\ast \in \mathbb{R}^{d}$ & hidden state \\ 
$\bm{q}^t \in \mathbb{R}^{d}$ & query at $t^{th}$ visit \\ \hline
$\bm{A}_b \in \mathbb{R}^{|\mathcal{C}_m| \times l}$ & adjacency matrix of bipartite graph  \\
$\bm{A}_e \in \mathbb{R}^{|\mathcal{C}_m| \times |\mathcal{C}_m|}$ & adjacency matrix of $G_e$\\ 
$\bm{A}_d \in \mathbb{R}^{|\mathcal{C}_m| \times |\mathcal{C}_m|}$ & adjacency matrix of $G_d$\\ \hline
$\bm{M}_b \in \mathbb{R}^{|\mathcal{C}_m| \times d}$ & Memory Bank (MB) \\ 
$\bm{M}_d^t \in \mathbb{R}^{|t-1| \times (d+|\mathcal{C}_m|)}$ & Dynamic Memory (DM) \\ 
$\bm{M}_{d,k}^t \in \mathbb{R}^{|t-1| \times d}$ & Keys in DM \\ 
$\bm{M}_{d,v}^t \in \mathbb{R}^{|t-1| \times |\mathcal{C}_m|}$ & Values in DM \\ \hline
$\bm{a}_c^t \in \mathbb{R}^{|\mathcal{C}_m|}$ & content-attention weight \\ 
$\bm{a}_s^t \in \mathbb{R}^{|t-1|}$ & temporal-attention weight \\ 
$\bm{a}_m^t \in \mathbb{R}^{|\mathcal{C}_m|}$ & history medication distribution \\ \hline
$\bm{o}_\ast^t \in \mathbb{R}^{d}$ & memory output \\ \hline
$\hat{\bm{y}}_t \in \mathbb{R}^{|\mathcal{C}_m|}$ & multi-label predictions at $t^{th}$ visit                                          \\ 
$\hat{Y}$           & recommended medication set \\ 
$Y$     & ground truth of medication set \\ \bottomrule[1pt]
\end{tabular}
\end{table}


\subsection{The \mname}
\begin{figure*}[ht]
\centerline{\includegraphics[width=\textwidth]{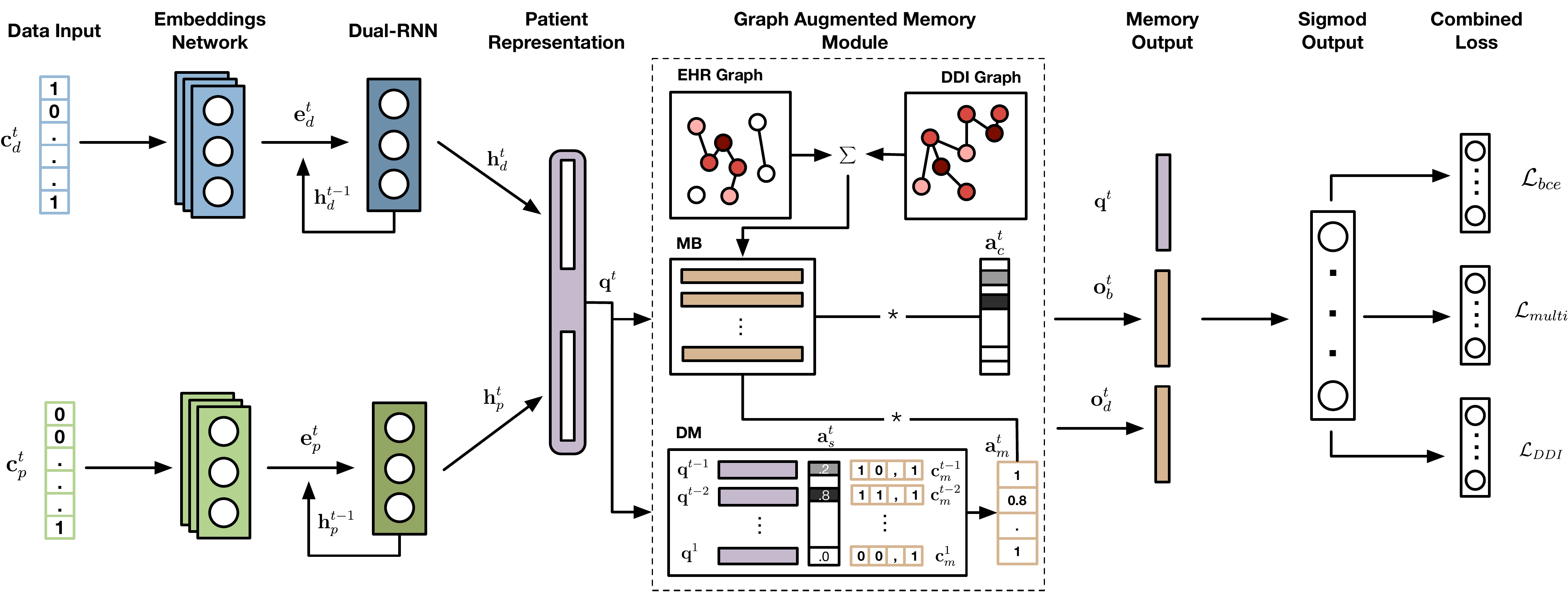}}
\caption{The \mname: At current $t^{th}$ visit, the multi-hot input $\bm{c}_d^t, \bm{c}_p^t$ are input into Embedding Networks to generate embedding $\bm{e}_d^t, \bm{e}_p^t$ using Eq.~\ref{eq:medical_embedding}. Then Dual-RNN generates current hidden states $\bm{h}_d^t, \bm{h}_p^t$ by accepting both embeddings from Embeddings Network and longitudinal hidden state $\bm{h}^{t-1}$ of RNN denoted by return arrow described in Eq.~\ref{eq:dual_rnn}. We use concatenated $\bm{h}_d^t, \bm{h}_p^t$ as query $\bm{q}^t$ (a.k.a. patient representation) in Eq.~\ref{eq:queries} to output $\bm{o}_b^t$ by reading from Memory Bank (MB) $\bm{M}_b$ in Eq.~\ref{eq:retrieve_mb} generated from late-fusion based multiple knowledge graph in Eq.\ref{eq:gen_m_2},~\ref{eq:gen_m_3}. Meantime, the Dynamic Memory (DM) stores key-value form history information along time by Eq.~\ref{eq:gen_md} and can be used to generate $\bm{o}_d^t$ in Eq.~\ref{eq:retrieve_mb}. Finally, query and memory outputs are concatenated in Eq.~\ref{eq:output} to make recommendation. In training phase, combined loss Eq.~\ref{eq:loss} is optimized to find optimal model parameters.}
\label{fig:framework}
\end{figure*}
As illustrated in Fig.~\ref{fig:framework}, \mname includes the following components: a medical embedding module, a patient representation module, and a graph augmented memory module. Next we will first introduce these modules and then provide details of training and inference of \mname.

\subsubsection{Medical Embeddings Module}\label{sec:med_code}
As mentioned before, a visit $\bm{x}_t$ consists of $[\bm{c}_d^t, \bm{c}_p^t, \bm{c}_m^t]$ where each of $\bm{c}_\ast^t$ is a multi-hot vector at the $t^{th}$ visit. The multi-hot vector $\bm{c}_\ast^t$ is binary encoded showing the existence of each medical codes recorded at the $t^{th}$ visit. Like~\cite{choi2016retain} used a linear embedding of the input vector,  we derive medical embeddings for $\bm{c}_d^t, \bm{c}_p^t$ separately at the $t^{th}$ visit as follows:
\begin{equation} \label{eq:medical_embedding}
\bm{e}_\ast^t = \bm{W}_{\ast, e} \bm{c}_\ast^t
\end{equation}
where $\bm{W}_{\ast, e} \in \mathbb{R}^{|\mathcal{C}_\ast| \times d}$ is the embedding matrix to learn. Thus a visit $\bm{x}_t$ is transformed to $\hat{\bm{x}}_t = [\bm{e}_d^t, \bm{e}_p^t, \bm{c}_m^t]$. 

\subsubsection{Patient Representation Module}\label{sec:tri-rnn}
To enable personalized medication recommendation which is tailored using patient EHR data, we design a Dual-RNN to learn patient representations from multimodal EHR data where each RNN encodes only one type of medical codes. The reason is that it is quite possible for a clinical visit to have missing modality (e.g. only diagnosis modality without procedure). Because of that, we model diagnosis and procedure modalities separately using two RNNs. Formally, for each input vector in transformed clinical history $[\hat{\bm{x}_1}, \hat{\bm{x}}_2, \cdots, \hat{\bm{x}}_{t}]$, we retrieve $\bm{e}_m, \bm{e}_p$ and utilize RNN to encode visit-level diagnosis and procedure embeddings respectively as follows:
\begin{eqnarray} 
&\bm{h}_d^t &= RNN_d(\bm{e}_d^1,\bm{e}_d^2,\cdots,\bm{e}_d^{t}) \nonumber \\
&\bm{h}_p^t &= RNN_p(\bm{e}_p^1,\bm{e}_p^2,\cdots,\bm{e}_p^{t})
\label{eq:dual_rnn}
\end{eqnarray}
Thus, the RNNs accept all patient history visit medical embeddings $\{\bm{e}_\ast^{t^\prime}\} (t^\prime \leq t)$  to produce hidden states $\bm{h}_\ast^t $ for further generating query (a.k.a. patient representation) in Eq.~\ref{eq:queries}.

\subsubsection{Graph Augmented Memory Module}\label{sec:graph_mod}
To leverage drug knowledge, we construct a graph augmented memory module that not only embeds and stores the EHR graph and the DDI graph as facts in Memory Bank (MB), but also inserts patient history to Dynamic Memory (DM) key-value form to fully capture the information from different views. 
Inspired by ~\cite{DBLP:journals/corr/WestonCB14}, four memory components \textbf{I}, \textbf{G}, \textbf{O}, \textbf{R} are proposed which mimics the architecture of modern computer in some way:

\begin{itemize}

\item \textbf{I: Input memory representation} converts inputs into query for memory reading. Here we can use hidden states from Dual-RNN to generate query as follows:
\begin{eqnarray}
\bm{q}^t = f([\bm{h}_d^t, \bm{h}_p^t]) \label{eq:queries}
\end{eqnarray}
where we concatenate hidden diagnosis state $\bm{h}_d^t$ and procedure state $\bm{h_p}^t$ as the input patient health state. $f(\cdot)$ is the transform function which projects hidden states to query and is implemented as single hidden layer fully connected neural network.

\item \textbf{G: Generalization} is the process of generating and updating the memory representation. We design the memory module by storing graph augmented memory representation as facts in Memory Bank (MB) and insert patient history to Dynamic Memory (DM) as key-value pairs to fully capture the information from different view.

For Memory Bank (MB) $\bm{M}_b$, two adjacency matrices $\bm{A}_e, \bm{A}_d$ are used. Following the GCN procedure ~\cite{kipf2016semi}, each $\bm{A}_\ast$ is preprocessed as follows:
\begin{eqnarray} 
\tilde{\bm{A}}_\ast = \tilde{\bm{D}}^{-\frac{1}{2}}(\bm{A}_\ast+\bm{I})\tilde{\bm{D}}^{-\frac{1}{2}}
\label{eq:gen_m_2}
\end{eqnarray}
where $\tilde{\bm{D}}$ is a diagonal matrix such that $\tilde{\bm{D}}_{ii}=\sum_j \bm{A}_{ij}$ and $\bm{I}$ are identity matrices.

Then we applied a two-layer GCN on each graph to learn improved embeddings on drug combination usage and DDIs respectively. The output $\bm{M}_b$ is generated as a weighted sum of the two graph embeddings.
\begin{eqnarray} 
&\bm{Z}_1 &= \tilde{\bm{A}}_{e} \text{tanh}(\tilde{\bm{A}}_{e}\bm{W}_{e1})\bm{W}_1 \nonumber \\
&\bm{Z}_2 &= \tilde{\bm{A}}_{d} \text{tanh}(\tilde{\bm{A}}_{d}\bm{W}_{e2})\bm{W}_2 \nonumber \\
&\bm{M}_b &= \bm{Z}_1 - \beta \bm{Z}_2
\label{eq:gen_m_3}
\end{eqnarray}
where $\bm{W}_{e1}$, $\bm{W}_{e2} \in \mathbb{R}^{|\mathcal{C}_m|\times d}$ are  medication embeddings from EHR graph  and DDI graph (each contains $|\mathcal{C}_m|$ number of d-dimensional vectors), $\bm{W}_1$, $\bm{W}_2 \in \mathbb{R}^{d \times d}$ are hidden weight parameter matrices. All $\bm{W}_\ast$ are updated during training phase. Then, graph node embeddings $\bm{Z}_1$, $\bm{Z}_2 \in \mathbb{R}^{|\mathcal{C}_m| \times d}$ are generated using GCN. Finally we combine different node embeddings as Memory Bank $\bm{M}_b \in \mathbb{R}^{|\mathcal{C}_m| \times d}$ where $\beta$ is a weighting variable to fuse different knowledge graphs. 

For Dynamic Memory (DM) $\bm{M}_d^t$, the combined patient  $\{\bm{q}^{t^\prime}\} (t^\prime < t)$ (the keys) associated with corresponding multi-hot medication vector $\{\bm{c}_m^{t^\prime}\}$ (the values) are inserted into DM as key-value pairs. This kind of design provides a way to locate most similar patient representation over time and retrieve the proper weighted medications set. Specifically, we can incrementally insert key-value pair after each visit step and treat $\bm{M}_d^t$ as a vectorized indexable dictionary as follows:
\begin{equation}
	\bm{M}_d^{t} = \{\bm{q}^{t^\prime} \colon \bm{c}_m^{t^\prime} \}_1^{t-1}
\label{eq:gen_md}
\end{equation}
where $\bm{M}_d^t$ is empty when $t=1$. For clarity, we use $\bm{M}_{d,k}^t = [\bm{q}^1;\bm{q}^2;\cdots;\bm{q}^{t-1}] \in \mathbb{R}^{|t-1| \times d}$ to denote the key vectors and $\bm{M}_{d,v}^t = [\bm{c}_m^1;\bm{c}_m^2;\cdots;\bm{c}_m^{t-1}] \in \mathbb{R}^{|t-1| \times |\mathcal{C}_m|}$ to denote the value vectors at $t^{th}$ visit.

\item \textbf{O: Output memory representation} produces outputs $\bm{o}_b^t$ and $\bm{o}_d^t$ given the patient representation $\bm{q}^t$ (the query) and the current memory state $\bm{M}_b, \bm{M}_d^t$. Here, we apply attention based reading procedure to retrieve most relevant information with respect to query $\bm{q}^t$ as outputs $\bm{o}_b^t, \bm{o}_d^t$ as follows:

\begin{eqnarray}
&\bm{o}_b^t &= \bm{M}_b^\intercal \overbrace{\text{Softmax}(\bm{M}_b \bm{q}^t)}^{\bm{a}_c^t} \nonumber \\
&\bm{o}_d^t &= \bm{M}_b^\intercal \overbrace{({\bm{M}_{d,v}^t})^\intercal \underbrace{\text{Softmax}(\bm{M}_{d,k}^t \bm{q}^t)}_{\bm{a}_s^t}}^{\bm{a}_m^t}
\label{eq:retrieve_mb}
\end{eqnarray}

where $\bm{o}_b^t \in \mathbb{R}^d$ is directly retrieved using content-attention $\bm{a}_c^t$ based on similarity between patient representation (query) and facts in $\bm{M}_b$. 

For $\bm{o}_d^t \in \mathbb{R}^d$, it firstly considers similar patient representation from patient history records $\bm{M}_{d,k}^t$ with temporal-attention $\bm{a}_s^t$. Then $\bm{a}_s^t$ is utilized to generate history medication distribution $\bm{a}_m^t$ by weighted sum of history multi-hot medication in $\bm{M}_{d,v}^t$. Finally, we can get $\bm{o}_d^t$ by further retrieved information from $\bm{M}_b$ using $\bm{a}_m^t$ from temporal aspect.

In addition, the attention based reading procedure makes the model differentiable so that it can be updated end-to-end using back propagation.

\item \textbf{R: Response} is the final step to utilize patient representation and memory output to predict the multi-label medication as follows:
\begin{eqnarray} 
\hat{\bm{y}}_{t} =  \sigma ([\bm{q}^t,\bm{o}_b^t, \bm{o}_d^t])
\label{eq:output}
\end{eqnarray}
where $\sigma$ is the sigmoid function.
\end{itemize}

\subsection{Training and Inference}
In the training phase, we need to find the optimal parameters including embedding matrix $\bm{W}_{e1}, \bm{W}_{e2},\bm{W}_{\ast,e}$, weight parameter matrix $\bm{W}_1, \bm{W}_2$ in GCN,  hidden weight in $f(\cdot), \text{RNN}$ as \textbf{auxiliary model parameter $\theta$}. We introduce the combined loss in order to find an optimal balance between recommendation accuracy and safety. At the end of the part, training algorithm will be given.\\

\noindent\textbf{Multi-label Prediction Loss (MLL)} Since the medication combination recommendation can be seen as sequential multi-label prediction, we combine two commonly used multi-label loss functions, namely, the binary cross entropy loss $\mathcal{L}_{bce}$ and the multi-label margin loss $\mathcal{L}_{multi}$. We use $\mathcal{L}_{multi}$ since it optimizes to make the predicted probability of ground truth labels has at least 1 margin larger than others. Thus, threshold value in Equation.~\ref{eq:inference} is easier to be fixed. 
\begin{eqnarray}
&\mathcal{L}_{bce} =  - \sum_t^T \sum_i \bm{y}_i^t \log \sigma(\hat{\bm{y}}_i^t)
+ (1 - \bm{y}_i^t) \log (1 - \sigma(\hat{\bm{y}}_i^t)) \nonumber \\
&\mathcal{L}_{multi} = \sum_t^T \sum_i^{|\mathcal{C}_m|} \sum_j^{|\hat{Y}^t|} \frac{\max(0, 1 - (\hat{\bm{y}}_t[\hat{Y}_j^t] - \hat{\bm{y}}_t[i]))}{L} \nonumber \\
&\mathcal{L}_p = \pi[0] L_{bce} + \pi[1]L_{multi}
\label{eq:multi-label loss}
\end{eqnarray}
where $\hat{\bm{y}}[i], \hat{\bm{y}}_i^t$ means the value at $i^{th}$ coordinate at $t^{th}$ visit, $\hat{\bm{y}}^t[\hat{Y}_j^t]$ means $j^{th}$ predicted label indexed by predicted label set $\hat{Y}^t$ at $t^{th}$ visit and $\bm{\pi}[\cdot]$ are the mixture weights ($\pi[0],\pi[1]\ge 0$, $\pi[0] + \pi[1] =1$).\\

\noindent\textbf{DDI Loss (DDI)} is designed to control DDIs in the recommendation.
\begin{eqnarray*}
\mathcal{L}_{DDI} = \sum_t^T \sum_{i,j} (\bm{A}_{d} \odot (\hat{\bm{y}}_{t}^\intercal \hat{\bm{y}}_{t}))[i, j] 
\end{eqnarray*}
where every element in $\hat{\bm{y}}_{t}^\intercal \hat{\bm{y}}_{t} \in \mathbb{R}^{N\times N}$ gives the pair-wise probability of predicted result. $\odot$ is the element-wise product. Intuitively, for two memory representation $i$,$j$, if $i$ $j$ combined to induce a DDI, then $\bm{A}_d[i,j] = 1 $. Thus large pair-wise DDI probability will yield large $\mathcal{L}_{DDI}$. \\

\noindent\textbf{Combined Loss functions}
When training, the accuracy and DDI Rate often increase together. The reason is that drug-drug interactions also exist in real EHR data (ground truth medication set $Y$). Thus both the incorrectly predicted medications and correctly predicted medications may increase the DDI Rate. To achieve the accurate model with low DDI Rate $s$ we need to find the balance between MLL and DDI. Inspired by Simulated Annealing~\cite{kirkpatrick1983optimization}, we can transform between NRL and MLL with a certain probability as follows:
\begin{eqnarray} 
\mathcal{L} = \begin{cases}
\mathcal{L}_p & if\text{ }s^\prime \leq s\\
\mathcal{L}_{DDI}, \text{with prob. }  p=\exp(-\frac{s^\prime - s}{Temp})  & if\text{ }s^\prime > s\\
\mathcal{L}_{p}, \text{with prob. }  p=1 - \exp(-\frac{s^\prime - s}{Temp})  & if\text{ }s^\prime > s
\end{cases}
\label{eq:loss}
\end{eqnarray}
on one hand, there will be high probability to use $\mathcal{L}_{DDI}$ when the DDI Rate $s^\prime$ of recommended medication set calculated in this step is larger than the expected DDI Rate $s$. On the other hand, decay rate $\epsilon$ applied on temperature $Temp \leftarrow \epsilon Temp$ makes $p$ low when model becomes stable along training time. Current DDI Rate $s^\prime$ can be calculated using DDI Rate Equation (see Metrics in Experiments section below) without sum across all test samples. The idea to use combined loss like simulated annealing form helps the model find best combination of parameters to demonstrate effectiveness and safety in the meantime.
In inference phase, thank to MLL, if the correctly predicted labels have at least 1 margin larger than others we can fix threshold value as 0.5. Then, the predicted label set corresponds to:
\begin{eqnarray} 
\hat{Y}_t = \{\hat{\bm{y}}_t^j | \hat{\bm{y}}_t^j > 0.5, 1 \leq j \leq ||\mathcal{C}_m||\}.
\label{eq:inference}
\end{eqnarray}
The training algorithm is detailed as follows. 
\begin{algorithm}[ht]
\caption{Training  \mname}
\label{alg:SMN}\small
\begin{algorithmic}
\REQUIRE Training set $\bm{R}$, training epoches $N$, mixture weight $\bm{\pi}$ in Eq.~\ref{eq:multi-label loss}, expected DDI Rate $s$, initial temperature $Temp$ and weight decay $\epsilon$ in Eq.~\ref{eq:loss}; 
\STATE Calculate adjacency matrix $\bm{A}_\ast$;
\STATE Using uniform distribution to initialize auxiliary model parameters $\theta \sim U(-1, 1)$;
\STATE Obtain Memory Bank $\bm{M}_b$ using Eq.~\ref{eq:gen_m_2},~\ref{eq:gen_m_3};

\FOR {$i = 1$ to $N \ast |\bm{R}|$}
	\STATE Sample a patient $\bm{P}=[\bm{x}_1, \bm{x}_2, \cdots, \bm{x}_{T_i}]$ from $\bm{R}$; 
    \STATE Reset Dynamic Memory $\bm{M}_d$;
	\FOR {$t = 1$ to $T_i$}
	\STATE Obtain medical embeddings $\bm{e}_d^t, \bm{e}_p^t$ in Eq.~\ref{eq:medical_embedding};
    \STATE Obtain Dual-RNN $\bm{h}_d^t, \bm{h}_d^t$ in Eq.~\ref{eq:dual_rnn};
	\STATE Generate patient representation $\bm{q}^t$ in Eq.~\ref{eq:queries};
    \STATE Read from $\bm{M}_b$ and $\bm{M}_d^t$ using attention weight $\bm{a}_c^t, \bm{a}_s^t, \bm{a}_m^t$ and generate memory outputs $\bm{o}_b^t,\bm{o}_d^t$ in Eq.~\ref{eq:retrieve_mb}; 
    \STATE Calculate medication prediction $\hat{\bm{y}_{t}}$ using Eq.~\ref{eq:output};
    \STATE Generate $\bm{M}_d^{t+1}$ by inserting $(\bm{q}^t, \bm{c}_m^t)$ into $\bm{M}_d^{t}$ in Eq.~\ref{eq:gen_md};
\ENDFOR
\STATE Evaluate and obtain DDI Rate $s^\prime$ of current patient;
\STATE Update $\theta$ by optimizing loss in Eq.~\ref{eq:loss} and decay $Temp \leftarrow \epsilon Temp$;
\ENDFOR
\end{algorithmic}
\end{algorithm}

\section{Experiments}

\subsection{Experimental Setup}
We evaluate \mname\footnote{https://github.com/sjy1203/GAMENet} model by comparing against other baselines on recommendation accuracy and successful avoidance of DDI. 
All methods are implemented in PyTorch~\cite{pytorch} and trained on an Ubuntu 16.04 with 8GB memory and Nvidia 1080 GPU.

\subsubsection{Data Source}
We used EHR data from MIMIC-III~\cite{johnson2016mimic}. Here we select a cohort where patients have more than one visit. In practice, if we use all the drug codes in an EMR record, the medication
set can be very large, each day in hospital, the doctor can prescribe
several types of medications for the patient. Hence, we choose the set of medications prescribed by doctors during the first 24-hour as the first 24-hour is often the most critical time for patients to obtain correct treatment quickly. In addition, we used DDI knowledge from TWOSIDES dataset~\cite{tatonetti2012data}. In this work, we keep the Top-40 severity DDI types and transform the drug coding from NDC to ATC Third Level for integrating with MIMIC-III.   
The statistics of the datasets are summarized in Table~\ref{tab:data}.
\begin{table}[ht]
\centering
\caption{Statistics of the Data }
\label{tab:data}
\begin{tabular}{l|l}
\toprule[1pt]
\# patients     &  6,350      \\ 
\# clinical events   &   15,016            \\ 
\# diagnosis   &   1,958  \\ 
\# procedure   &   1,426          \\ 
\# medication   &  145             \\ \hline
avg \# of visits   &   2.36  \\
avg \# of diagnosis   &   10.51  \\ 
avg \# of procedure   &   3.84          \\ 
avg \# of medication   &  8.80            \\ \hline
\# medication in DDI knowledge base   &    123           \\ 
\# DDI types in knowledge base    &          40    \\ \bottomrule[1pt]
\end{tabular}
\end{table}

\subsubsection{Baselines}
We consider the following baseline algorithms.
\begin{itemize}
\item \textbf{Nearest} will simply recommend the same combination medications at previous visit for current visit (i.e., $\hat{Y}_t = Y_{t-1}$)
\item \textbf{Logistic Regression (LR)} is a logistic regression with L2 regularization. Here we represent the input data by sum of one-hot vector. Binary relevance technique ~\cite{luaces2012binary} is used to handle multi-label output. 
\item \textbf{Leap} \cite{zhang2017leap} is an instance-based medication combination recommendation method. 
\item \textbf{RETAIN} \cite{choi2016retain} can provide sequential prediction of medication combination based on
a two-level neural attention model that detects influential past visits and significant clinical variables within those visits.
\item \textbf{DMNC} \cite{le2018dual} is a recent work of medication combination prediction via memory augmented neural network based on differentiable neural computers (DNC)~\cite{graves2016hybrid}. 
\end{itemize}

\subsubsection{Metrics}
To measure the prediction accuracy, we used Jaccard Similarity Score (Jaccard), Average F1 (F1) and Precision Recall AUC (PRAUC).
Jaccard is defined as the size of the intersection divided by the size of the union of ground truth medications $Y_t^{(k)}$ and predicted medications $\hat{Y}_t^{(k)}$.
\begin{eqnarray}
\text{Jaccard} = \frac{1}{\sum_k^N \sum_t^{T_k} 1}\sum_k^N \sum_t^{T_k} \frac{|Y_t^{(k)} \cap \hat{Y}_t^{(k)}|}{|Y_t^{(k)} \cup \hat{Y}_t^{(k)}|}\nonumber
\end{eqnarray}
where $N$ is the number of patients in test set and $T_k$ is the number of visits of the $k^{th}$ patient.
Average Precision (Avg-P) and Average Recall (Avg-R), and F1 are defined as:
\begin{eqnarray}
&\text{Avg-P}_t^{(k)} &=  \frac{|Y_t^{(k)} \cap \hat{Y}_t^{(k)}|}{|Y_t^{(k)}|} ,\ 
\text{Avg-R}_t^{(k)} =  \frac{|Y_t^{(k)} \cap \hat{Y}_t^{(k)}|}{|\hat{Y}_t^{(k)}|} \nonumber \\
&\text{F1} =& \frac{1}{\sum_k^N \sum_t^{T_k} 1}\sum_k^N \sum_t^{T_k} \frac{2 \times \text{Avg-P}_t^{(k)} \times \text{Avg-R}_t^{(k)}}{\text{Avg-P}_t^{(k)} + \text{Avg-R}_t^{(k)}} \nonumber
\end{eqnarray}
where $t$ means the $t^{th}$ visit and $k$ means the $k^{th}$ patient in test dataset.

To measure medication safety, we define DDI Rate as percentage of medication recommendation that contain DDIs.
\begin{eqnarray}
\text{\small DDI Rate} = \frac{\sum_k^N \sum_t^{T_k} \sum_{i, j} |\{(c_i, c_j) \in \hat{Y}_t^{(k)}| (c_i, c_j) \in \mathcal{E}_d \}| }{\sum_k^N\sum_t^{T_k} \sum_{i, j} 1}\nonumber
\end{eqnarray}
where the set will count each medication pair $(c_i, c_j)$ in recommendation set $\hat{Y}$ if the pair belongs to edge set $\mathcal{E}_d$ of the DDI graph. Here $N$ is the size of test dataset and $T_k$ is the number of visits of the $k^{th}$ patient.

The relative DDI Rate ($\triangle$ DDI Rate \%) is defined as the percentage of DDI rate change compared to DDI rate in EHR test dataset given DDI rate of the algorithm:
\begin{equation}
\triangle\text{ DDI Rate \%} = \frac{\text{DDI Rate - DDI Rate (EHR)}}{\text{DDI Rate (EHR)}}\nonumber
\end{equation}

\subsubsection{Evaluation Strategies}
We randomly divide the dataset into training, validation and testing set in a $2/3 : 1/6 : 1/6$ ratio. For LR, we use the grid search technique over typical range of hyper-parameter to search the best hyperparameter values which result in L2 norm penalty with weight as 1.1. For our methods, the hyperparameters are adjusted on evaluation set which result in expected DDI  Rate $s$ as 0.05, initial temperature $Temp$ as 0.5, weight decay $\epsilon$ as 0.85 and mixture weights $\bm{\pi} = [0.9, 0.1]$. For all deep learning based methods, we choose a gated recurrent
unit (GRU)~\cite{cho2014properties} as the implementation of RNN and utilize dropout~\cite{srivastava2014dropout} with probability of an element to be zeroed as 0.4 on the output of embeddings. The embedding size and hidden layer size for GRU is set as 64 and 64 respectively, word and memory size for DMNC model is 64 and 16 which is the same as~\cite{le2018dual}. Training is done through Adam~\cite{DBLP:journals/corr/KingmaB14} at learning rate 0.0002. We fix the best model on evaluation set within 40 epochs and report the performance in test set. 

\subsection{Results}

\begin{table*}[ht]
\caption{Performance Comparison of Different Methods. Note that the base DDI rate in EHR test data is  0.0777. } 
\label{tab:exp_res}
\centering
\scalebox{0.95}[0.95]{
\begin{tabular}{l|c|c|c|c|c|c|c}
\toprule[1pt]
 Methods &  DDI Rate & $\triangle$ DDI Rate \% & Jaccard & PR-AUC  & F1 &  Avg \# of Med.  & \# of parameters \\ \hline
 Nearest & 0.0791    & $+\ 1.80\%$ & 0.3911   & 0.3805   & 0.5465 & 14.77 & -\\ \hline
 LR      & 0.0786    & $+\ 1.16\%$ & 0.4075   & 0.6716   & 0.5658 & 11.42 & -\\ \hline
 Leap    & \bf{0.0532}  & \bf{$-$\ 31.53\%}  & 0.3844   & 0.5501 & 0.5410   & 14.42 & 436,884\\ \hline
 RETAIN  & 0.0797  & $+\ 2.57\%$  & 0.4168   & 0.6620   & 0.5781 & 16.68 & 289,490\\ \hline
 DMNC    & 0.0949  & $+\ 22.14\%$ & 0.4343   & 0.6856   & 0.5934 & 20.00 & 527,979\\ \hline
 \small \mname(w/o DDI)    & 0.0853  & $+\ 9.78\%$  & 0.4484    & 0.6878   & 0.6059 & 15.13 & 452,434\\ \hline
 \mname   & \bf{0.0749}  & \bf{$-$\ 3.60\%}  & \bf{0.4509}   & \bf{0.6904}    & \bf{0.6081} & 14.02 & 452,434\\\bottomrule[1pt]
\end{tabular}
}
\end{table*}

\begin{table*}[ht]
\caption{Example Recommended Medications for a Patient with Two Visits. Here "unseen" indicates the medications predicted but are not in ground truth, while "missed" refers to the medications that are in the ground truth but are not predicted.}
\label{tab:exp_case_study}
\centering
\scalebox{0.76}[0.76]{
\begin{tabular}{l|l|l}
\toprule[1pt]
\bf Diagnosis & \bf Methods & \bf Recommended Medication Combination \\ 
\midrule
\multirow{6}{*}{
\begin{tabular}[c]{@{}l@{}} 
\textbf{1st Visit}:\\ Malignant neoplasm \\of brain\\ Hyperlipidmia\\ Gout
\end{tabular}
} 
& Ground Truth & \small N02B, A01A, A02B, A06A, B05C, A12C, C07A, C02D, N02A, B01A, C10A, J01D, N03A, A04A, H04A \\ \cline{2-3}
& Nearest & \small 0 correct  + 15 missed \\ \cline{2-3} 
& LR      & \small 3 correct (N02B, A01A, A06A) + 12 missed \\ \cline{2-3} 
& Leap    & \small 8 correct (N02B, A02B, A06A, A12C, C07A, B01A, C10A, A04A)  + 7 missed \\ \cline{2-3} 
& RETAIN  & \small 0 correct  + 15 missed \\ \cline{2-3} 
& DMNC    & \small 12 correct (N02B, A01A, A02B, A06A, B05C, A12C, C07A, C02D, N02A, B01A, C10A, N03A) + 6 unseen  + 3 missed \\ \cline{2-3} 
& \mname    & \small 11 correct (N02B, A01A, A02B, A06A, B05C, A12C, C07A, C02D, B01A, N03A, A04A) + 4 missed 
\\ \hline
\multirow{6}{*}{
\begin{tabular}[c]{@{}l@{}} 
\textbf{2nd Visit}:\\ Malignant neoplasm \\of brain\\ Cerebral Edema\\ Hypercholesterolemia\\ Gout
\end{tabular}
}
& Ground Truth & \small N02B, A01A, A02B, A06A, B05C, A12C, C07A, C02D, N02A, B01A, J01D, N03A, N05A, A04A \\ \cline{2-3}
& Nearest & \small 13 correct (N02B, A01A, A02B, A06A, B05C, A12C, C07A, C02D, N02A, B01A, J01D, N03A, A04A) + 2 unseen + 1 missed \\ \cline{2-3} 
& LR      & \small 3 correct (N02B, A01A, A06A)  + 11 missed \\ \cline{2-3} 
& Leap    & \small 7 correct (N02B, A01A, A02B, A06A, B05C, A12C, B01A) + 2 unseen + 7 missed \\ \cline{2-3} 
& RETAIN  & \small 10 correct (N02B, A01A, A02B, A06A, B05C, A12C, C07A, N02A, B01A, N03A) + 5 unseen  + 4 missed \\ \cline{2-3} 
& DMNC    & \small 12 correct (N02B, A01A, A02B, A06A, B05C, A12C, C07A, C02D, N02A, B01A, N03A, A04A) + 7 unseen   + 2 missed \\ \cline{2-3} 
& \mname    & \small  13 correct (N02B, A01A, A02B, A06A, B05C, A12C, C07A, C02D, N02A, B01A, J01D, N03A, A04A) + 1 unseen + 1 missed \\ 
\bottomrule[1.2pt]
\end{tabular}}
\end{table*}

Table~\ref{tab:exp_res} compares the performance on accuracy and safety issue. Results show \mname
has the highest score among all baselines  with respect to Jaccard, PR-AUC, and F1. 

As for the baseline models,  Nearest and LR achieved about 4\% lower score compared to \mname in terms of Jaccard and F1. The Nearest method also gives us the clue that the visit is highly important for the medications combination recommendation task. For both methods, the DDI rates are very close to the base DDI rate in the EHR data. This implies without knowledge guidance that it will be hard to remove DDIs that already exist in clinical practice.
For deep learning baselines, instance-based method Leap achieved lower performance than those temporal models such as RETAIN and DMNC, which confirmed the important of temporal information in patient past EHRs.

On the other hand, for longitudinal methods such as RETAIN and DMNC, they both achieve higher scores on Jaccard, PRAUC, and F1 compared with others. DMNC however recommends a large bunch of medication combination set which may be one reason that lead to high DDI Rate. 

For our methods, we compare the \mname and its variant \mname(w/o DDI). Without DDI knowledge, \mname(w/o DDI) is also better than other methods which shows the overall framework does work. With DDI knowledge, both the performance and DDI rate are improved. The result is statistically significant using two-tailed t-test after ten runs of these two methods.

\subsection{Case Study}

We choose a patient from test dataset based on the consideration of demonstrating the model effect on harder cases: there are diagnoses and medications change among visits. As shown in Table.~\ref{tab:exp_case_study}, the patient has 3 diagnoses for the $1^{st}$ visit and two extra diagnoses Cerabral Edema, Hypercholesterolemia for the $2^{nd}$ visit. The ground truth medications prescribed by doctors and recommended medications by different methods are listed in the table. Overall, \mname performs the best with 11 correct, 13 correct medications  for two visit respectively, only missed 4 and 1 medications and wrongly predict 1 (unseen) medication for $2^{nd}$ visit. For Nearest and RETAIN methods, they lack the ability to recommend medication combination for $1^{st}$ visit. DMNC tries to recommend more medications which result in more wrongly predicted medications than other methods. To mention that, all methods except LR and \mname will recommend the combination of N02B (Analgesics and Antipyretics) and C10A (Lipid-modifying Agents), which can lead to harmful side effect such as Myoma. This harmful combination also existed in ground truth of the patient's at $1^{st}$ visit. For the $2^{nd}$ visit, C10A is removed from ground truth medications set, which may indicate doctors also try to correct their decision. Another pair of medications A01A (Stomatological Preparations) and N03A (Antiepileptic Drugs) exists in the ground truth of both visits. Their combined use could cause allergic bronchitis. Most methods including Nearest, RETAIN, DMNC recommend them. \mname also recommends them due to the trade off between effectiveness and safety.

\section{Conclusion}
In this work, we presented \mname, an end-to-end deep learning model
that aims to generate effective and safe recommendations of medication combinations via memory networks whose memory bank is augmented by integrated drug usage and DDI graphs as well as dynamic memory based on patient history.
Experimental results on real-world EHR showed that \mname outperformed all baselines in effectiveness measures, and achieved $3.60\%$ DDI rate reduction from existing EHR data. As we noticed the trade-off between effectiveness and safety measures, a possibly rewarding avenue of future research is to simultaneous recommend medication replacements that share the same indications of the harmful drugs but will not induce adverse DDIs.  

\section*{Acknowledgment}

This work was supported by Peking University Medicine Seed Fund for Interdisciplinary Research, the National Science Foundation, award IIS-1418511 and CCF-1533768, the National Institute of Health award 1R01MD011682-01 and R56HL138415. We would also like to thank Tianyi Tong, Shenda Hong and Yao Wang for helpful discussions.

\bibliographystyle{aaai}
\bibliography{sample}

\end{document}